\documentclass[english, aip, reprint]{revtex4-2}
\usepackage[T1]{fontenc}
\usepackage{amstext}
\usepackage{amssymb}
\usepackage{amsmath}
\usepackage{babel}
\usepackage{graphicx}
\usepackage[caption=false]{subfig}

\usepackage{xcolor}
\usepackage[normalem]{ulem}

\begin{document}

\title{Understanding the stochastic dynamics of sequential decision-making processes: A path-integral analysis of multi-armed bandits }

\author{Bo Li}
\email{libo2021@hit.edu.cn}
\affiliation{School of Science, Harbin Institute of Technology (Shenzhen), Shenzhen, 518055, China}

\affiliation{Non-linearity and Complexity Research Group, Aston University, Birmingham, B4 7ET, United Kingdom}

\author{Chi Ho Yeung}
\email{chyeung@eduhk.hk}
\affiliation{Department of Science and Environmental Studies, The Education University
of Hong Kong, 10 Lo Ping Road, Tai Po, Hong Kong}

\begin{abstract}
The multi-armed bandit (MAB) model is one of the most classical models to study decision-making in an uncertain environment.
In this model, a player chooses one of $K$ possible arms of a bandit machine to play at each time step, where the corresponding arm returns a random reward to the player, potentially from a specific unknown distribution. The target of the player is to collect as many rewards as possible during the process.
Despite its simplicity, the MAB model offers an excellent playground for studying the trade-off between exploration versus exploitation and designing effective algorithms for sequential decision-making under uncertainty.
Although many asymptotically optimal algorithms have been established, the finite-time behaviors of the stochastic dynamics of the MAB model appear much more challenging to analyze, due to the intertwine between the decision-making and the rewards being collected.
In this paper, we employ techniques in statistical physics to analyze the MAB model, which facilitates the characterization of the distribution of cumulative regrets at a finite short time, the central quantity of interest in an MAB algorithm, as well as the intricate dynamical behaviors of the model.
Our analytical results, in good agreement with simulations, point to the emergence of an interesting multimodal regret distribution, with large regrets resulting from excess exploitation of sub-optimal arms due to an initial unlucky output from the optimal one.
\end{abstract}

\maketitle

\begin{quotation}
Making decisions in an uncertain environment is a common and challenging task faced by human beings, other animals and intelligent machines. 
In such tasks, the decision-maker usually has several possible options, where the outcome of each option is unknown and stochastic. We investigate a scenario where the decision-maker has a limited budget to test the options, from which he/she gets rewards from the actions and simultaneously gains knowledge of the options with increasing confidence, which forms the basis of future decision-making among those repeated options.
Such a scenario constitutes a complex dynamical process where the decision-making and the collected rewards are intertwined, especially in finite time.
To unveil the nature of such complex dynamical processes, we apply methods from statistical physics to analyze the multi-armed bandit model, which is a prototypical mathematical model capturing the above characteristics of decision-making. 
The analytical distribution of the outcome of the decision-making process in finite time agrees well with simulations, and depicts the origin of rare events where outcomes are much better or worse than expected can occur.
\end{quotation}

\section{Introduction}\label{sec:intro}

Decision-making and optimization in uncertain environments are ubiquitous tasks faced by human beings, other natural creatures and intelligent machines. For example, animals have to decide on which unknown environment patches to search for food resources, which is crucial for their survival~\cite{Viswanathan2011}.
Optimization of repeated dynamical decisions is also essential in a large group of complex systems, such as the planning of path for swarm robots~\cite{wang2022expert}, repeated prisoner dilemma in social networks~\cite{qiu2023continuous} and collective decision making in Parrondo’s games~\cite{lai2020social}.
Such tasks are also common in a pandemic situation, where one needs to decide on whether to adopt new life-saving medicines or vaccines with limited supporting evidence, or to research these medications to understand their pros and cons more thoroughly before decisions are made~\cite{Villar2015}; in terms of testing, to decide which areas or groups to prioritize testing resources without a comprehensive understanding of the infection mechanism when testing capacity is limited~\cite{GrushkaCohen2020}. 
Sticking to familiar good solutions can fully exploit the knowledge from past experiences but may lose the opportunity to discover possible better solutions, which is sub-optimal in the long run. On the other extreme, always exploring new territories can increase the chance to search for the best solutions but the excessive exploration may incur a high cost. The root of such decision-making tasks is to strike a balance  between exploration and exploitation~\cite{Sutton2018}.

The multi-armed bandit problem is one of the most classical models to address this issue~\cite{Bubeck2012,Slivkins2019}. In this model, the bandit machine consists of $K$ independent arms; each arm $k\in\{1,2,...,K\}$ will return a reward $x_{k}\in\mathcal{\mathbb{R}}$ drawn from an unknown probability distribution $P_{k}(x_{k})$ once it is pulled by the player. 
The task of the player is to choose an arm $a^{t}\in\{1,2,...,K\}$ at time $t$ based on the historical outcomes of the rewards $\{ x^{\tau}_{a^{\tau}} \}_{\tau=0}^{t-1}$, in order to maximize the cumulative reward $R=\sum_{t=0}^{T}x^{t}_{a^{t}}$ for a time period $T$. 
We denote the expected reward of arm $k$ as $\mu_{k}=\mathbb{E}_{P_{k}}[x_{k}]$, and define the best arm $k^{*}$ as the arm having the largest expected reward $k^{*}=\text{arg}\max_{k}\mu_{k}$. Maximizing the cumulative rewards is equivalent to minimizing the cumulative regrets $r = \sum_{t=0}^{T} (\mu_{*} - x^{t}_{a^{t}})$ with $\mu_{*}=\mu_{k^{*}}$ as the mean reward of the best arm, since $\mu_{*}$ is a constant. 
From the player's perspective, the regret is not accessible as $\mu_{*}$ is unknown, so the player will use the cumulative reward $R$ as the objective function in practice. However, the cumulative regret $r$ is more useful for theoretically assessing the player's strategy in hindsight.

Despite the simplicity of the MAB model, it captures the essential characteristics of decision-making under uncertainty, in that the rewards returned by each arm are probabilistic and that the player has to devise a strategy to make decisions based on the noisy observations. 
There exists a lot of research works on the multi-armed bandit models in the statistics and applied mathematics literature, most of which aim to bound the expected cumulative regrets in the long run. Notably, optimal asymptotic bounds are established for certain policies to play the bandit machines under some appropriate conditions of the reward distributions, which states that as the time period $T$ increases, the expected regrets $\mathbb{E}[r]$ only grows as $O(\log T)$, provided that the gap is sufficiently large between the optimal and sub-optimal arms~\cite{Lai1985}. 
The MAB model and the corresponding optimal strategies also receive a lot of attentions in the fields of reinforcement learning and the optimization of black-box functions, since the theoretical understanding established on the former is relevant to the latter~\cite{Sutton2018,Shahriari2016}. 

Most existing theoretical studies focus on the expected behaviors of MAB. However, even when a strategy is optimal in the average sense, it may still incur high regrets in some realizations of the processes due to fluctuations of the rewards or the actions of the player (if they are stochastic)~\cite{Audibert2009, Salomon2011}. 
In some applications, safety is also of concern in addition to collecting more rewards on average, which may require a more conservative strategy in general. To this end, many risk-averse bandit algorithms were proposed~\cite{Sani2012, Galichet2013}. Most existing research efforts along this line addressed this issue by augmenting or replacing the expected rewards with some risk-aware measures (such as variance), and focused on devising algorithms for the new problems.

A more in-depth understanding of the fine-grained probabilistic characteristics of the deviations from the average behaviors including rare events is important for such purposes, but it remains much less explored. Some developments in such directions were made recently by mapping some bandit algorithms onto stochastic differential equations and considering the asymptotic limit~\cite{Wager2021, Fan2021}.
From another perspective, the continual interactions between the player and the bandit machine constitute an interesting stochastic dynamical system, which may exhibit rich emergent behaviors.

In this study, we aim to investigate the probabilistic nature of the regret distribution and shed light on the MAB model as a complex stochastic dynamical system through the lens of the large deviation analysis combined with a path-integral formalism from statistical physics~\cite{Touchette2009,Hertz2016}, and to provide insights on decision-making processes under uncertainty. 
Compared to many existing studies which are based on high-probability bounds and use the asymptotic limit of a long time (i.e. a large number of trials), our analysis can provide more detailed and explicit characteristics of MABs in less explored regimes more difficult to be analyzed, including those rare events with a small probability in a finite time. 
We emphasize that our target is not to introduce new bandit algorithms or improve existing methods. 

\section{The Model}

In this section, we describe the canonical multi-armed bandit model and some famous bandit algorithms for solving the corresponding decision-making problems.

\subsection{Problem formulation}
The essential elements of the MAB model have been briefly introduced in Sec.~\ref{sec:intro}. Here we provide more details of the model and clarify the notations that will be used. We consider a fixed time period $0\leq t \leq T$ of the game. At time $t=0$, the player pulls each arm $k\in \{ 1,2, ... ,K \}$ once to warm up the system, and receives the corresponding random reward $x^{0}_{k} \sim P_{k}(x)$ for each arm $k$. 
This can be considered as a homogeneous initial exploration step. At each time step $t \in \{1,2,...,T\}$, the player selects only one arm $a^{t} \in \{ 1,2,...,K \}$ among all $K$ options based on the historical observations of the model, and receives a random reward $x^{t}_{a^{t}}$ from the corresponding arm.

We denote $n^{t}_{k}$ as the number of times that arm $k$ has been pulled up to time $t$, and $s^{t}_{k}$ as the total rewards received from arm $k$ up to time $t$. The quantities $n^{t}_{k}$ and $s^{t}_{k}$ satisfy
\begin{align}
n_{k}^{0} & = 1, \label{eq:n0_def} \\
s_{k}^{0} & = x^{0}_{k}, \label{eq:s0_def} \\
n_{k}^{t} & = \sum_{\tau=1}^{t}\delta(a^{\tau},k)+n_{k}^{0}, \label{eq:n_def} \\
s_{k}^{t} & = \sum_{\tau=1}^{t}x_{a^{\tau}}^{\tau}\delta(a^{\tau},k)+s_{k}^{0}, \label{eq:s_def}
\end{align}
where $\delta(\cdot, \cdot)$ is the Kronecker delta function. That is, the action $a^{\tau}$ at time $\tau$ has a contribution to $n^{t}_{k}$ and $s^{t}_{k}$ only if $a^{\tau} = k$ when $\tau \leq t$.
We also adopt the short-hand notations of the vector along the time dimension $s^{0:T}_{k} := [s^{0}_{k}, s^{1}_{k}, ..., s^{T}_{k}]$, and the vector along the arm dimension $\boldsymbol{s}^{t} := [s^{t}_{1}, s^{t}_{2}, ..., s^{t}_{K}]$. Under these notations, the cumulative reward is $R = \sum_{k} s_{k}^{T}$ and the cumulative regret is $r = (T+K) \mu_{*} - \sum_{k} s_{k}^{T}$.

The policy (or algorithm) of a MAB model specifies a strategy of choosing an arm $k = a^{t+1}$ at time $t+1$ based on previous actions $\{ a^{\tau} \}_{\tau \leq t}$ and observations $\{ x^{\tau}_{a^{\tau}} \}_{\tau \leq t}$. Due to the non-interacting nature of the basic MAB model, it is convenient to consider the sum of historical choices and payoffs $\{ n^{t}_{k'}, s^{t}_{k'} \}_{\forall k'}$ to determine which arm to pull at time $t$. In this case, the action $a^{t+1}$ is a function of sum of these historical statistics as $a^{t+1}( \boldsymbol{n}^{t}, \boldsymbol{s}^{t})$. 
Since the parameters of the distributions of the arms are unknown, we need to estimate them based on historical observations. For instance, after pulling the arm at time $t$, the expected reward of arm $k$ can be estimated as 
\begin{equation}
\hat{\mu}_{k}^{t} = \frac{s_{k}^{t}}{n_{k}^{t}}, \label{eq:mu_hat_def}
\end{equation}
which will be used by the player for determining which arm to pull in the future.

\subsection{Bandit Algorithms}
A simple strategy for optimizing outcomes from the MAB model is to perform the following two-stage operation (i) allocate a fixed number of trials to each arm $k$ to estimate the expected reward $\hat{\mu}_{k}$ by Eq.~(\ref{eq:mu_hat_def}) (this is a pure exploration phase); (ii) identify the estimated best arm $\hat{k}^{*} = \text{arg}\max_{k}\hat{\mu}_{k}$ and keep pulling it in the future steps (this is a pure exploitation phase). 
In order to precisely estimate the arm parameters for the benefit of the second stage, a substantial amount of resources is needed to explore all arms, resulting in low rewards being collected from the non-optimal arms in the first stage. On the other hand, if fewer resources are spent in the first stage for exploration, then the player has a higher risk of incorrectly identifying the best arm. 
Many real-life problems are solved by this policy, e.g., in clinical trials, it is usual to perform a predefined amount of experimental trials to choose one medicine or vaccine among several possible options and then deploy the chosen one widely. 
Such a policy can be sub-optimal for the purpose of maximizing the cumulative rewards. Many research efforts are devoted to finding a better policy (or algorithm) for MAB which strikes a balance between exploration and exploitation. 

\subsubsection{$\epsilon$-greedy Algorithm}
An alternative strategy is to perform exploration and exploitation stochastically in an online fashion. In the $\epsilon$-greedy algorithm~\cite{Sutton2018}, at each time step, the player either chooses the estimated best arm to pull (with probability $1-\epsilon$), or selects a random arm to explore (with probability $\epsilon$). The probability $\epsilon$ controls the amount of exploration, which can vary over time. 

\subsubsection{Softmax Algorithm}
Another stochastic algorithm is the softmax method~\cite{Sutton2018}, which picks an arm at time $t+1$ according to a Boltzmann distribution
\begin{equation}
p_{k}^{t+1} = \frac{ e^{ \beta \hat{\mu}_{k}^{t}} }{ \sum_{j=1}^{K} e^{\beta \hat{\mu}_{j}^{t}} },
\end{equation}
where $\beta$ is the inverse temperature parameter controlling the amount of exploration. For a small $\beta$, it tends to explore more uniformly among different arms, while for a large $\beta$, it tends to exploit the estimated best arm. The inverse temperature can also vary over time.

\subsection{UCB Algorithm}
Despite the simplicity of the above-mentioned stochastic algorithms, they either yield sub-optimal total rewards or require carefully tuning the parameters. 
Another important class of bandit algorithms, the upper confidence bounds (UCB) algorithms, stem from more solid theoretical properties and achieve optimal asymptotic total rewards on average~\cite{Auer2002, Bubeck2012, Slivkins2019}. It starts by computing the upper confidence bound (UCB) of the sample mean estimation for each arm $k$ as
\begin{align}
B_{k}^{t} & = \frac{s_{k}^{t}}{n_{k}^{t}} + b^{t} \frac{1}{\sqrt{n_{k}^{t}}}, \label{eq:UCB_def1}
\end{align}
where the parameter $b^{t}$ controls the confidence level, which can vary over time. Since the sample standard deviation of a certain random variable is proportional to $1/\sqrt{n}$ after $n$ independent measurements, Eq.~(\ref{eq:UCB_def1}) represents an upper confidence bound for reward $x_{k}$ after $n^{t}_{k}$ measurements on arm $k$. 
The UCB algorithm proceeds by selecting the arm corresponding to the highest UCB index $a^{t+1} = \text{arg}\max_{k} B_{k}^{t}$, which is a principle of ``optimism in the face of uncertainty'' (by comparing the best possibilities of all arms in a certain confidence level). 

The two terms of the UCB index in Eq.~(\ref{eq:UCB_def1}) represent the effects of exploitation and exploration, respectively. The second term of Eq.~(\ref{eq:UCB_def1}) encourages exploration because if an arm $k$ has a low $n_{k}^{t}$ (low exploration of the arm up to time $t$), then the second term is large for arm $k$, which increases its UCB index and the chance that arm $k$ is being pulled.

If the rewards have a bounded range, e.g., $0 \leq x_{k} \leq 1$, it is sufficient to achieve the optimal asymptotic regret bound of $O(\log(T))$ by setting the tuning parameter $b^{t-1}$ in Eq.~(\ref{eq:UCB_def1}) as $b^{t} = c\sqrt{\log(K+t)}$, where $c$ is a parameter tuning the level of exploration; notice that $K+t$ is the total number of arm pulls until time $t$, because each of the $K$ arms was pulled once at the initial time. 

\subsection{MABs as Stochastic Dynamical Systems}
Despite the simplicity of the problem formulation, the MABs represent a fairly complex decision-making process from the dynamical system perspective, as the random rewards being collected will impact the decisions in future rounds, and this will in turn impact the estimate of rewards given by individual arms and hence the reward given by the bandit machine in the future. 

To disentangle such complex dynamics, we will adopt methods from statistical physics on stochastic dynamical systems for analyzing MABs. To this end, we consider the probabilistic evolution of the action $\{ a^{t} \}_{1 \leq t \leq T}$ in the form of, given by
\begin{align}
& B_{k}^{t}(s_{k}^{t},n_{k}^{t}) = \frac{s_{k}^{t}}{n_{k}^{t}} + c \sqrt{\frac{\log(K + t)}{n_{k}^{t}}}, \\
& P(a^{t+1}=k) = \frac{e^{\beta B_{k}^{t}(s_{k}^{t},n_{k}^{t})}}{\sum_{j}e^{\beta B_{j}^{t}(s_{j}^{t},n_{j}^{t})}}, \qquad t \geq 0, \label{eq:P_of_a_def}
\end{align}
which is a combination of the softmax and UCB strategies. In the infinite $\beta$ (zero temperature) limit, Eq.~(\ref{eq:P_of_a_def}) reduces to the UCB algorithm. On the other hand, it becomes the traditional softmax algorithm when $c=0$. 

In this work, we specify the arm reward distributions as Gaussian distributions,
\begin{equation}
P(x_{k}^{t})=\mathcal{N}(x_{k}^{t}|\mu_{k},\sigma_{k}^{2}). \label{eq:P_of_x_def}
\end{equation}

\section{Statistical-physics Analysis}
We now proceed to analyze the stochastic dynamical process of MABs by adopting methods from statistical physics. The quantity of interest is the probability distribution of the cumulative regret
\begin{align}
P(r) & = \bigg\langle \delta \bigg[ r - \big( (T+K)\mu_{*} - \sum_{k}s_{k}^{T} \big) \bigg] \bigg\rangle_{a^{1:T}, \boldsymbol{x}^{0:T}}, \nonumber \\
& = \int\prod_{t=0}^{T-1}\mathrm{d}\boldsymbol{x}^{t}P(\boldsymbol{x}^{t})\prod_{t=1}^{T} \sum_{a^{t}} P(a^{t}) \nonumber \\
& \qquad \times \delta\big(r+\sum_{k}s_{k}^{T}-(T+K)\mu_{*}\big),  \label{eq:P_of_r_def}
\end{align}
where the average is taken over the dynamical variables $a^{1:T}$ and $\boldsymbol{x}^{0:T}$ with respect to the distributions given in Eq.~(\ref{eq:P_of_a_def}) and Eq.~(\ref{eq:P_of_x_def}).

\subsection{Path-integral Formalism}
We note that in Eq.~(\ref{eq:P_of_a_def}), the variables $s_{k}^{t}$ and $n_{k}^{t}$ also depend on the historical trajectories $a^{1:t}$ and $\boldsymbol{x}^{0:t}$ through Eq.~(\ref{eq:n_def}) and Eq.~(\ref{eq:s_def}), making the average in Eq.~(\ref{eq:P_of_r_def}) highly non-trivial. Borrowing methods from statistical physics, we consider $\{r, \boldsymbol{n}^{0:T},  \boldsymbol{s}^{0:T} \}$ as statistical fields and adopt the path-integral formalism for computing this average. 

We first express the delta function in Eq.~(\ref{eq:P_of_r_def}) by its Fourier representation 
\begin{equation}
\delta\big(r+\sum_{k}s_{k}^{T}-(T+K)\mu_{*}\big) = \int \frac{d\hat{r}}{2\pi} e^{-\mathrm{i} \hat{r}\big(r+\sum_{k}s_{k}^{T}-(T+K)\mu_{*}\big)},
\end{equation}
and insert the Fourier representation of unities for  $\{ s_{k}^{t},n_{k}^{t} \}$ through the definition in Eq.~(\ref{eq:n0_def})-Eq.~(\ref{eq:s_def}) to Eq.~(\ref{eq:P_of_r_def})
\begin{align}
1 & = \int \frac{\mathrm{d} \hat{n}_{k}^{0} \mathrm{d} n_{k}^{0}}{2\pi} \exp\big[ -\mathrm{i} \hat{n}_{k}^{0} \big( n_{k}^{0} - 1 \big) \big], \\
1 & = \int \frac{\mathrm{d} \hat{s}_{k}^{0} \mathrm{d} s_{k}^{0}}{2\pi} \exp\big[ -\mathrm{i} \hat{s}_{k}^{0} \big( s_{k}^{0} - x_{k}^{0} \big) \big], \\
1 & = \int \frac{\mathrm{d} \hat{n}_{k}^{t} \mathrm{d} n_{k}^{t}}{2\pi} \exp\bigg[ -\mathrm{i} \hat{n}_{k}^{t} \bigg( n_{k}^{t}-1-\sum_{\tau=1}^{t}\delta(a^{\tau},k) \bigg) \bigg], \\
1 & = \int \frac{\mathrm{d} \hat{s}_{k}^{t} \mathrm{d} s_{k}^{t}}{2\pi} \exp\bigg[ -\mathrm{i} \hat{s}_{k}^{t} \bigg( s_{k}^{t}-x_{k}^{0}-\sum_{\tau=1}^{t}x_{k}^{\tau}\delta(a^{\tau},k) \bigg) \bigg],
\end{align}
after which Eq.~(\ref{eq:P_of_r_def}) becomes 
\begin{align}
P(r) & = \int\prod_{t=0}^{T}\mathrm{d}\boldsymbol{x}^{t}P(\boldsymbol{x}^{t})\int\prod_{k=1}^{K}\prod_{t=0}^{T}\frac{\mathrm{d}\hat{s}_{k}^{t}\mathrm{d}s_{k}^{t}}{2\pi}\frac{\mathrm{d}\hat{n}_{k}^{t}\mathrm{d}n_{k}^{t}}{2\pi}\frac{\mathrm{d}\hat{r}}{2\pi} \nonumber \\
& \sum_{\{a^{t}\}_{t=1}^{T}} \prod_{t=1}^{T} \bigg[ e^{\beta B_{a^{t}}^{t-1}(s_{a^{t}}^{t-1},n_{a^{t}}^{t-1})} / \sum_{k'=1}^{K}e^{\beta B_{k'}^{t-1}(s_{k'}^{t-1},n_{k'}^{t-1})} \bigg] \nonumber \\
& \times \exp \bigg\{ -\mathrm{i} \hat{r}\big(r+\sum_{k}s_{k}^{T}-(T+K)\mu_{*}\big) \bigg\} \nonumber \\
& \times \exp\bigg\{-\mathrm{i}\sum_{k=1}^{K}\bigg[\hat{s}_{k}^{0}\big(s_{k}^{0}-x_{k}^{0}\big)+\hat{n}_{k}^{0}\big(n_{k}^{0}-1\big)\bigg]\bigg\} \nonumber \\
& \times \exp\bigg\{-\mathrm{i}\sum_{k=1}^{K} \sum_{t=1}^{T}\hat{s}_{k}^{t}\bigg(s_{k}^{t}-x_{k}^{0}-\sum_{\tau=1}^{t}x_{k}^{\tau}\delta(a^{\tau},k)\bigg) \bigg\} \nonumber \\
& \times \exp\bigg\{ -\mathrm{i}\sum_{k=1}^{K} \sum_{t=1}^{T}\hat{n}_{k}^{t}\bigg(n_{k}^{t}-1-\sum_{\tau=1}^{t}\delta(a^{\tau},k)\bigg) \bigg\}. \label{eq:Pr_path_int_1st}
\end{align}
In the second line of Eq.~(\ref{eq:Pr_path_int_1st}), the variables $\{ s_{k}^{t},n_{k}^{t} \}$ are stochastic fields which do not explicitly depend on $a^{1:t}$ and $\boldsymbol{x}^{0:t}$. Instead, such dependencies are expressed through their coupling to the conjugate fields $\{ \hat{s}_{k}^{t}, \hat{n}_{k}^{t} \}$. 

Now the average over the disorder variables $\boldsymbol{x}^{0:T}$ in Eq.~(\ref{eq:Pr_path_int_1st}) can be easily perform by using the identity $\int e^{-\mathrm{i}xh}\mathcal{N}(x|\mu,\sigma^{2})dx=e^{-\frac{1}{2}\sigma^{2}h^{2}-\mathrm{i}\mu h}$, such that the distribution of regret $P(r)$ has the form of
\begin{align}
P(r) = \int\prod_{k=1}^{K}\prod_{t=0}^{T}\frac{\mathrm{d}\hat{s}_{k}^{t} \mathrm{d}s_{k}^{t}}{2\pi}\frac{\mathrm{d}\hat{n}_{k}^{t} \mathrm{d} n_{k}^{t}}{2\pi}\frac{d\hat{r}}{2\pi}e^{-\Phi}, \label{eq:Pr_of_Phi}
\end{align}
where $\Phi$ is a stochastic action defined as
\begin{align}
\Phi & = \mathrm{i}\hat{r}\bigg(r+\sum_{k}s_{k}^{T}-(T+K)\mu_{*}\bigg) \nonumber \\
& +\mathrm{i}\sum_{k}\sum_{t=0}^{T}\big(\hat{s}_{k}^{t}s_{k}^{t}+\hat{n}_{k}^{t}(n_{k}^{t}-1)\big)  \nonumber \\
& + \frac{1}{2}\sum_{k}\sigma_{k}^{2}\sum_{t,t'=0}^{T}\hat{s}_{k}^{t}\hat{s}_{k}^{t'}n_{k}^{\min(t,t')} -\mathrm{i}\sum_{k}\mu_{k}\sum_{t=0}^{T}\hat{s}_{k}^{t}n_{k}^{t} \nonumber \\
& - \sum_{t=1}^{T}\bigg\{\log\sum_{k}\exp\bigg[\beta B_{k}^{t-1}(s_{k}^{t-1},n_{k}^{t-1}) +\mathrm{i}\sum_{\tau=t}^{T}\hat{n}_{k}^{\tau} \bigg] \nonumber \\
& \qquad - \log\sum_{k}e^{\beta B_{k}^{t-1}(s_{k}^{t-1},n_{k}^{t-1})}\bigg\}. \label{eq:Phi_def}
\end{align}
See Appendix~\ref{sec:appendix_a} for details of the calculations. 
The action function $\Phi$ is reminiscent of a thermodynamic potential, which expresses the stochastic dynamical system through the order parameters $\{ s_{k}^{t}, n_{k}^{t}, \hat{s}_{k}^{t}, \hat{n}_{k}^{t} \} $ of all time steps. Hence, Eq.~(\ref{eq:Pr_of_Phi}) is a discrete-time path integral.

\subsection{Small Noise Limit and Saddle-point Equations}\label{sec:saddle}
An exact expression of Eq.~(\ref{eq:Pr_of_Phi}) is difficult to obtain. 
Therefore, we need methods of approximation to simplify the formalism. To this end, we focus on the limit of low temperature and small arm noise $\beta \to \infty, \sigma_{k} \to 0$, which essentially consider the large deviation of the regret under the UCB algorithm~\cite{Grafke2019}. This is because when the noise is small, the regret $r$ has a high probability to be near its most probable value $r^{\text{mpv}}$; in this case a sufficiently strong departure of $r$ from $r^{\text{mpv}}$ become rare events, which is the focus of large deviation theory in terms of probability~\cite{Touchette2009, Grafke2019}. 
In the regimes of large deviations, the dynamical events leading to the regret departure from $r^{\text{mpv}}$ are usually dominated by a single typical trajectory~\cite{Grafke2019}, which simplifies the analysis.
Although the theory below is built on such a small noise limit, we expect some physical properties of the system still hold in the finite noise case, which is another interesting regime of the MAB problem.
We refer readers to Ref.~\cite{Grafke2019} for more in-depth discussions about the small noise limit and the large deviation.

In the small noise limit, for a particular value of regret $r$, the path integral in Eq.~(\ref{eq:Pr_of_Phi}) is dictated by the saddle point of the action function $\Phi$~\cite{Grafke2019}, which satisfies $\partial \Phi / \partial y = 0, y \in \{ s_{k}^{t}, n_{k}^{t}, \hat{s}_{k}^{t}, \hat{n}_{k}^{t}, \hat{r} \}$. 

To make the notations more compact, we define the following quantities
\begin{align}
B^{t}_{k} & = B^{t}_{k}(s^{t}_{k}, n^{t}_{k}), \\
B^{t}_{s,k} & = \frac{\partial B^{t}_{k}(s^{t}_{k}, n^{t}_{k})}{\partial s^{t}_{k}}, \\
B^{t}_{n, k} & = \frac{\partial B^{t}_{k}(s^{t}_{k}, n^{t}_{k})}{\partial n^{t}_{k}}, \\
\rho_{k}(\boldsymbol{v}) & = \frac{e^{v_{k}}}{\sum_{j=1}^{k} e^{v_{j}}}, \\
h_{k}^{t+1} & = \beta B_{k}^{t}(s_{k}^{t},n_{k}^{t}) + \mathrm{i}\sum_{\tau > t}^{T}\hat{n}_{k}^{\tau},
\end{align}
where $\rho_{k}(\cdot)$ is the softmax operator.

Then the saddle point equations admit the following expressions
\begin{align}
n_{k}^{0} & = 1,  \\
n_{k}^{t} & = 1+\sum_{\tau=1}^{t}\rho_{k}(\boldsymbol{h}^{\tau}),\quad1\leq t\leq T, \label{eq:nkt_forward_1} \\
s_{k}^{t} & = \mu_{k}n_{k}^{t} + \sigma_{k}^{2}\sum_{t'=0}^{T}(\mathrm{i}\hat{s}_{k}^{t'})n_{k}^{\min(t,t')}	,\quad0\leq t\leq T, \label{eq:skt_forward_1} \\
\mathrm{i}\hat{s}_{k}^{t} & = \beta B_{s,k}^{t}\big[\rho_{k}(\boldsymbol{h}^{t+1})-\rho_{k}(\beta \boldsymbol{B}^{t})\big],\; 0\leq t < T, \label{eq:iSkt_backward_1} \\
\mathrm{i}\hat{s}_{k}^{T} & = -\mathrm{i}\hat{r}, \\
\mathrm{i}\hat{n}_{k}^{t} & = \beta B_{n,k}^{t}\big[\rho_{k}(\boldsymbol{h}^{t+1})-\rho_{k}(\beta \boldsymbol{B}^{t})\big]+\mu_{k}\mathrm{i}\hat{s}_{k}^{t} \nonumber \\  
& \quad -\sigma_{k}^{2}\sum_{t'>t}^{T}\hat{s}_{k}^{t}\hat{s}_{k}^{t'}-\frac{1}{2}\sigma_{k}^{2}(\hat{s}_{k}^{t})^{2},\quad0\leq t < T, \label{eq:iNkt_backward_1} \\
\mathrm{i}\hat{n}_{k}^{T} & = \mu_{k}\mathrm{i}\hat{s}_{k}^{T}-\frac{1}{2}\sigma_{k}^{2}(\hat{s}_{k}^{T})^{2}, \\
r & = (T+K)\mu_{*}-\sum_{k}s_{k}^{T},
\end{align}
where the last equation is a constraint for the total reward $\sum_{k} s_{k}^{T}$ (we remind that $r$ is a pre-defined parameter instead of a dynamical variable). We remark that the conjugate fields $\{ \hat{s}_{k}^{t}, \hat{n}_{k}^{t}, \hat{r} \}$ are defined on the imaginary axis in the saddle point~\cite{Paga2017, Li2020sensitivity}.
The nonlinear saddle point equations are difficult to solve analytically, but can be solved numerically.

\subsection{Simplification in the Large $\beta$ Limit}\label{sec:sim_saddle}
Further simplification can be made by exploiting the limits of large $\beta$ and small arm noise $\sigma_k$. In particular, we will firstly send $\beta \to \infty$, and then consider $\sigma_k \to 0$. We remind that both limits are essential for taking the saddle point approximation.

In this limit of large $\beta$, the field $h_{k}^{t+1}$ is dominated by $\beta B_{k}^{t}$, so that Eq.~(\ref{eq:nkt_forward_1}) can be approximated as
\begin{equation}
n_{k}^{t} = 1+\sum_{\tau=1}^{t}\rho_{k}(\beta \boldsymbol{B}^{\tau - 1}),\quad1\leq t\leq T, \label{eq:nkt_forward_2}.
\end{equation}

To derive the dynamics of the conjugate order parameters $\{ \hat{s}_{k}^{t}, \hat{n}_{k}^{t} \}_{t<T}$ in this limit, we notice 
\begin{align}
\rho_{k}(\boldsymbol{h}^{t+1}) & \approx \rho_{k}(\beta \boldsymbol{B}^{t}) + \sum_{j=1}^{K} \left. \frac{\partial \rho_{k}(\boldsymbol{h}^{t+1})}{\partial h_j^{t+1}} \right|_{ \boldsymbol{h}^{t+1} = \beta \boldsymbol{B}^{t} } \cdot \mathrm{i}\sum_{\tau > t}^{T}\hat{n}_{j}^{\tau} \nonumber \\
& =: \rho_{k}(\beta \boldsymbol{B}^{t}) + \hat{\Delta}\rho^{t+1}(\beta \boldsymbol{B}^{t}) \cdot \sum_{\tau > t}^{T} \mathrm{i} \boldsymbol{n}^{\tau},
\end{align}
where the $K\times K$ matrix $\hat{\Delta}\rho^{t+1}(\beta \boldsymbol{B}^{t})$ has the element $\hat{\Delta}\rho^{t+1}(\beta \boldsymbol{B}^{t})_{kj} = \delta_{jk} \rho_{k}(\beta \boldsymbol{B}^{t}) - \rho_{k}(\beta \boldsymbol{B}^{t})\rho_{j}(\beta \boldsymbol{B}^{t})$. 

Under this approximation, the dynamical rules of Eq.~(\ref{eq:iSkt_backward_1}) and Eq.~(\ref{eq:iNkt_backward_1}) become
\begin{align}
\mathrm{i}\hat{s}_{k}^{t} & = \beta B_{s,k}^{t} \sum_{j} \hat{\Delta}\rho^{t+1}_{kj} \sum_{\tau > t}^{T}\mathrm{i}\hat{n}_{j}^{\tau}, \\ 
\mathrm{i}\hat{n}_{k}^{t} & = \beta B_{n,k}^{t} \sum_{j} \hat{\Delta}\rho^{t+1}_{kj} \sum_{\tau > t}^{T}\mathrm{i}\hat{n}_{j}^{\tau} +\mu_{k}\mathrm{i}\hat{s}_{k}^{t} \nonumber \\
& \quad + \sigma_{k}^{2}\sum_{t'>t}^{T} (\mathrm{i}\hat{s}_{k}^{t}) (\mathrm{i}\hat{s}_{k}^{t'})+\frac{1}{2}\sigma_{k}^{2}(\mathrm{i}\hat{s}_{k}^{t})^{2}, 
\end{align}
which admits a backward iteration form.

In such a large $\beta$ limit, the stochastic action $\Phi$ \textit{evaluated at the saddle point} has a simple expression
\begin{equation}
\Phi^{*}(r | \sigma_{k}) = \frac{1}{2}\sum_{k}\sigma_{k}^{2}\sum_{t,t'=0}^{T} (\mathrm{i}\hat{s}_{k}^{t}) (\mathrm{i}\hat{s}_{k}^{t'}) n_{k}^{\min(t,t')}.
\end{equation}

We notice that if the variances of all arms are rescaled by a common factor $\gamma$, i.e., $\sigma_{k}^{2} \to \gamma \sigma_{k}^{2}$, then the conjugate order parameters $\{ \hat{s}_{k}^{t}, \hat{n}_{k}^{t}, \hat{r} \}$ evaluated at the saddle point equations change accordingly as $\hat{s}_{k}^{t} \to \hat{s}_{k}^{t}/\gamma, \hat{n}_{k}^{t} \to \hat{n}_{k}^{t}/\gamma, \hat{r} \to \hat{r}/\gamma$, while the order parameters $\{ s_{k}^{t}, n_{k}^{t} \}$ remain unchanged; the stochastic action changes as $\Phi^{*} \to \Phi^{*}/\gamma$. 

If we set the variance of arm $k$ as $\sigma_{k}^{2} = \gamma \Tilde{\sigma}_{k}^2$, where  $\Tilde{\sigma}_{k}^2$ is a fixed parameter and $\gamma$ can be varied (common to all arms), then the cumulative regret $r$ follows a large deviation principle as
\begin{align}
P(r|\sigma_{k}^2 = \gamma \Tilde{\sigma}_{k}^2) & \propto \exp\bigg( - \frac{1}{\gamma} I(r; \Tilde{\sigma}_{k}^2) \bigg), \label{eq:Pr_LDP} \\
I(r | \Tilde{\sigma}_{k}^2) & \equiv \gamma \Phi^{*}(r | \sigma_{k}),
\end{align}
where $I(r | \Tilde{\sigma}_{k}^2)$ is the rate function governing the rareness of regret $r$, and the order parameters $\{ s_{k}^{*t}, n_{k}^{*t} \}$ at the saddle point dictates the most probable pathway leading to the regret $r$ being specified~\cite{Grafke2019}.

We remark that such a relation is derived in the limit $\beta \to \infty, \sigma_{k} \to 0$, but we expect some physical pictures can also be extended to cases with a finite noise strength. The approximation in Eq.~(\ref{eq:nkt_forward_2}) essentially neglects some fluctuation of the arm selection noise from the Boltzmann distribution in Eq.~(\ref{eq:P_of_a_def}), which requires that the fluctuation due to a finite $\beta$ in Eq.~(\ref{eq:P_of_a_def}) is much smaller than the fluctuation from finite arm variances $\sigma_{k}^{2}$. 
Therefore, the limit of $\beta \to \infty$ should be taken before the limit $\sigma_{k} \to 0$.

We also remark that the saddle point approach introduced above is only a leading-order approximation, and it is possible to derive higher-order corrections for a better accuracy~\cite{Hertz2016}.

\section{Results}
In this section, we report the theoretical results from the above statistical-physics analysis on MABs in different scenarios and corroborate them with numerical experiments. 

We consider a bandit machine with $K=3$ independent arms, where the reward of each arm $k$ follows a Gaussian distribution $x_{k} \sim \mathcal{N}(\mu_{k}, \sigma_{k}^{2})$. The arm distribution parameters are set to be $\boldsymbol{\mu} = [1,2,3], \boldsymbol{\sigma} = \sqrt{\gamma}[1, 1, 1]$, which corresponds to setting $\Tilde{\sigma}_{k} = 1$ in Eq.~(\ref{eq:Pr_LDP}). One technical difficulty is solving the highly nonlinear saddle point equations with $4K(T+1) + 1$ variables. Hence, we focus on systems with a relatively small number of time steps ($T\sim O(10)$), which already exhibits many interesting dynamical phenomena.

\subsection{The Stochastic Action}
To solve the saddle-point equations, we adopt an iteration method described in Appendix~\ref{sec:iteration_method} in detail. 
We found that in some parameter regimes, the saddle point equations admit multiple solutions, in which cases we retained the solution with the smallest stochastic action $\Phi^{*}$. 

\begin{figure}
    \centering
    \includegraphics[scale=1.05]{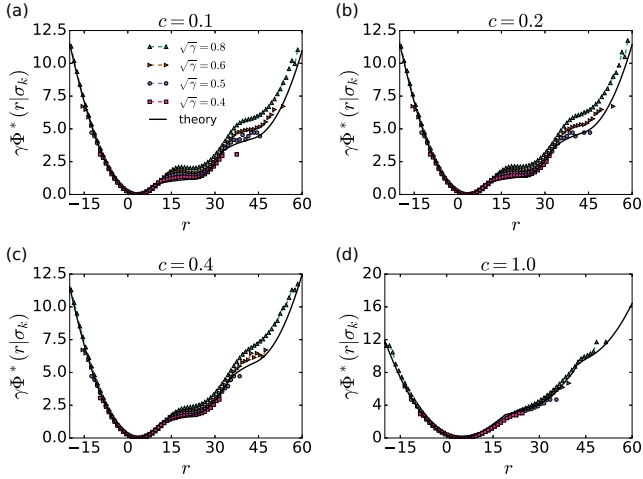}
    \caption{Rescaled stochastic action $\gamma \Phi^{*}(r | \sigma_{k})$ as a function of the cumulative regret $r$. The black lines represent the theoretical predictions, while the dashed lines with dots are obtained by numerical simulations of $1\times 10^9$ trials. The parameters are $K=3, T=20, \beta=10$. The theory in Sec.~\ref{sec:saddle} and \ref{sec:sim_saddle} predicts a universal rescaled stochastic action $\gamma \Phi^{*}(r|\sigma_{k}) = I(r|\Tilde{\sigma}_{k})$ in the low noise limit.}
    \label{fig:Phi_vs_r_vary_gamma_c}
\end{figure}

In Fig.~\ref{fig:Phi_vs_r_vary_gamma_c}, we sketch the resulting rescaled stochastic action $\gamma \Phi^{*}(r | \sigma^{k})$ for different arm variance magnitude $\gamma$ and different exploration parameter $c$, by fixing $\beta = 10$ and considering $T=20$. 
The rescaled potential $\gamma \Phi^{*}(r | \sigma^{k})$ predicted by the theory (black lines) exhibits a peculiar non-convex structure, comprised of 3 convex pieces (in general there are $K$ convex pieces for $K$-armed bandits). This indicates a multimodal structure of the regret distribution $P(r)$~\cite{Audibert2009}. 

We also compare the theoretical prediction of the stochastic action to numerical simulations, obtained by simulating the corresponding MAB for $1 \times 10^9$ trials. The stochastic action for MABs by simulation is defined as 
\begin{equation}
\Phi^{\text{sim}}(r) = - \log \hat{P}(r) - \min_{r} \big[ - \log \hat{P}(r) \big],
\end{equation}
where $\hat{P}(r)$ is the empirical density of the cumulative regret, and we have also subtracted $- \log \hat{P}(r)$ by its minimum. Under this definition, the minimal $\Phi^{\text{sim}}(r)$ is zero, which facilitates an easier comparison to the theory. 
We note that $\Phi^{\text{sim}}(r)$ loses the information about the precise value of $\hat{P}(r)$, but still carries the information of the relative strength between the probability densities of different regrets.  

The results shown in Fig.~\ref{fig:Phi_vs_r_vary_gamma_c} demonstrate a good match between theory and simulation for small and moderate arm variance level $\sqrt{\gamma}$. The difference between the two approaches becomes more prominent for large $\sqrt{\gamma}$, which is expected since the theory is developed in the small noise limit; but even in these cases, the qualitative trend of $\Phi^{\text{sim}}(r)$ still follows the theoretical prediction. 
In both theory and simulations, the stochastic action is asymmetric and right-skewed, which indicates that the MAB has a higher chance to be unlucky (the regret $r$ is larger than the most probable value $r^{\text{mpv}}$) than lucky (the regret $r$ is smaller than $r^{\text{mpv}}$). This is because being lucky mainly originates from the rarely good outcomes of many rounds from the optimal arm, but being unlucky originates from the rarely bad outcomes of a small number of rounds from the optimal arm at the beginning, which makes it not well explored.  The former occurs with a very small probability, while the latter occurs with a relatively higher probability.
The heavy right tail of the regret distribution indicates a noticeable probability of the total arm rewards being sub-optimal in some realizations in a short time, despite that the algorithm is optimal in the long run in the average sense~\cite{Audibert2009, Fan2021}.

\subsection{Effect of the Exploration Parameter $c$}
\begin{figure}
    \centering
    \includegraphics[scale=1.05]{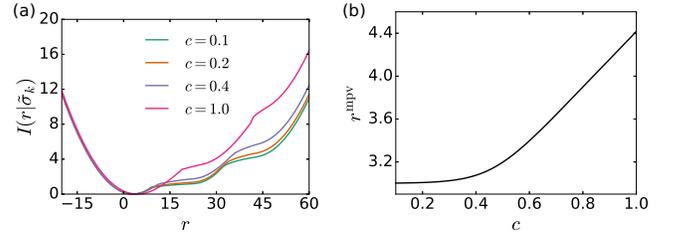}
    \caption{(a) The rate function $I(r|\Tilde{\sigma}_{k})$ predicted by the theory in the low noise limit $\beta \to \infty, \gamma \to 0$ as a function of regret $r$ for different exploration parameters $c$. (b) The most probable value of regret $r^{\text{mpv}} = \text{arg}\min_{r} I(r|\Tilde{\sigma}_{k})$ as a function of the exploration parameter $c$.}
    \label{fig:Phi_and_rmpv_vary_c}
\end{figure}

The noticeable probability for the system to have a high regret indicates that the system can be trapped into a state of excessive selections of sub-optimal arms (as will be shown in detail below), which is partly due to insufficient exploration. 
This effect can be alleviated by increasing the exploration parameter $c$, as shown in Fig.~\ref{fig:Phi_and_rmpv_vary_c}(a). It is observed that for a higher $c$, the rate function $I(r|\Tilde{\sigma}_{k})$ attains larger values in the unlucky regime (with large positive regrets), indicating a much lighter tail for $P(r)$ for $r\gg 0$ and hence a much lower probability for more unlucky events.

On the other hand, increasing the exploration parameter $c$ will increase the expected value of the regret, as the chance to explore sub-optimal arms also gets higher. Since it is not straightforward to measure the expected regret in theory since it involves the knowledge of the complete form of the rate function $I(r|\Tilde{\sigma}_{k})$ over the whole domain of $r$, we compute the most probable value of the regret $r^{\text{mpv}}$ for different $c$, as shown in Fig.~\ref{fig:Phi_and_rmpv_vary_c}(b). 
It is observed that $r^{\text{mpv}}$ increases very gently at the beginning, and starts to rise more significantly after $c \approx 0.4$. Therefore, the suppression of the right tail of the regret distribution $P(r)$ comes with a price of increasing the most probable regret. A risk-averse MAB algorithm needs to take these effects into account.

\subsection{Dominant Trajectories}
Another nice feature of the large deviation path-integral approach is that it can give rise to the dominant trajectories leading to a particular regret $r$. When the noise is small, the trajectory leading to the smallest action $\Phi^{*}$ at a particular $r$ dominates the probability density $P(r)$ in the path integral Eq.~(\ref{eq:Pr_of_Phi}). 
These dominant trajectories are also called the optimal paths or the instantons~\cite{Grafke2019}. Here, we use the characteristics of the optimal paths derived from the small noise limit to explain the dynamics behaviors of MABs with a finite noise strength, which is of more practical interest. 
We also consider a  small time window $T=20$, which already provides many valuable insights. 

\subsubsection{Lucky and slightly unlucky events: Conditioned on $r=-8$ and $r=6$}
\begin{figure}
    \centering
    \includegraphics[scale=1.05]{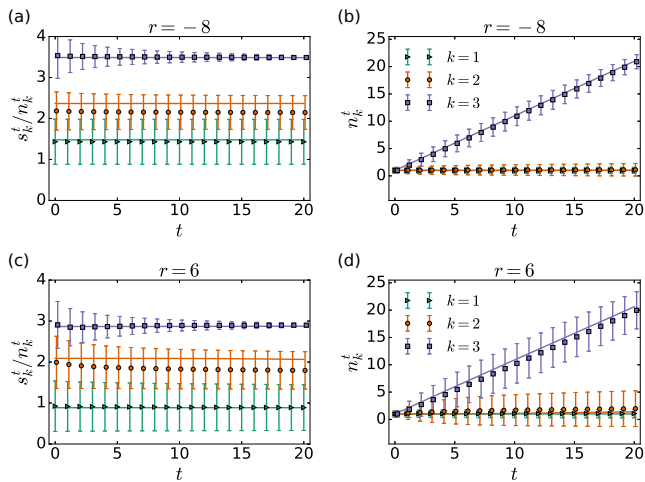}
    \caption{Dominant trajectories at a negative regret $r=-8$ (panel (a) and (b)) and a small positive regret $r=6$ (panel (c) and (d)), revealed by the dynamical evolution of $s_{k}^{t} / n_{k}^{t}$ and $n_{k}^{t}$. The lines represent the trajectories predicted by the theory. The dots with error bars are results from numerical simulations by simulating $1\times 10^9$ trials and keeping the trajectories with cumulative regret $r^{\text{sim}} \in [r, r+0.5)$; the error bars represent one standard deviation of the quantity measured in these trajectories. The parameters are $K=3, T=20, \beta=10, c=0.4, \sqrt{\gamma}=0.6$.}
    \label{fig:traj_small_r}
\end{figure}

In Fig.~\ref{fig:traj_small_r}, we consider the trajectories conditioned on a negative regret $r=-8$ and a small positive regret $r=6$, which is considered as a lucky case and a  slightly unlucky case. It is observed that in both cases, the optimal arm $k^{*} = 3$ is mostly chosen by the player after $t=0$, and the trajectory leading to a particular regret $r$ is due to homogeneous deviation of the realized reward $x_{k^{*}}^{t}$ at time $t$ from its expected value $\mu_{*}$ (as demonstrated in Fig.~\ref{fig:traj_small_r}(a) and (c)). That is, the deviation $x_{k^{*}}^{t} - \mu_{*}$ is almost the same for all $t>0$, which is akin to the ``fluid phase'' of independent random variables $\{ x_i \}$ conditioned by the value of their sum~\cite{Godrche2019}. 

The theoretical prediction and numerical simulations match very well, although the arm noise is not small.

\subsubsection{Unlucky events: Conditioned on $r=16$ and $r=24$}\label{sec:traj_r16_r24}
\begin{figure}
    \centering
    \includegraphics[scale=1.05]{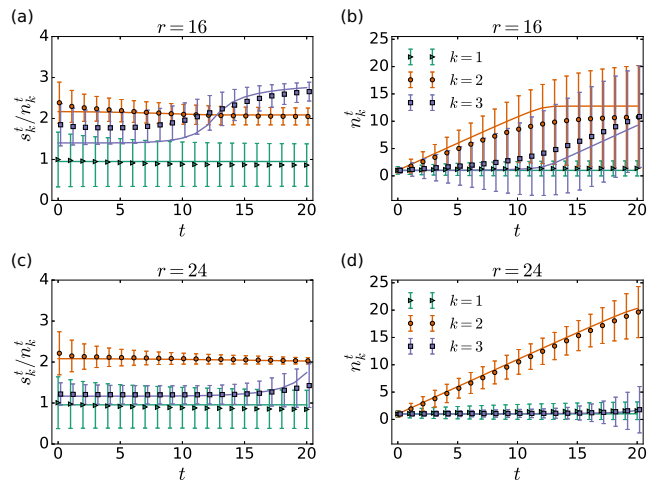}
    \caption{Dominant trajectories at regrets $r=16$ (panel (a) and (b)) and $r=24$ (panel (c) and (d)). The parameter setting is the same as those in Fig.~\ref{fig:traj_small_r}.}
    \label{fig:traj_medium_r}
\end{figure}

In Fig.~\ref{fig:traj_medium_r}, we consider the unlucky cases with moderate regrets with $r=16$ and $r=24$, which enters the second convex branch of the potential $\Phi^{*}(r)$ shown in Fig.~\ref{fig:Phi_vs_r_vary_gamma_c}(c). 
Interestingly, the typical trajectory leading to $r=16$ behaves as follows: the reward from the 3rd arm (the optimal arm) in the initial exploration stage is unluckily a bit lower than that from the 2nd arm (the sub-optimal arm), resulting in substantial exploitation of the 2nd arm in the subsequent steps. During this process, $n_{2}^{t}$ grows while $n_{3}^{t}$ remains unchanged. Up to a certain point, the exploration term $c\sqrt{\log(K+t)/n_{3}^{t}}$ on the 3rd arm becomes large enough for the UCB index $B_{3}^{t}$ of the 3rd arm to dominate that of the 2nd arm.
After then, the player starts to collect rewards from the 3rd arm and gradually realizes that it has a higher expected reward than the more exploited 2nd arm. 

Similarly, for $r=24$, the reward from the 3rd arm in the initial exploration step is also unluckily small, and it is even smaller than the case of $r=16$. Therefore, the 2nd arm is being pulled throughout the process. As the gap between the empirical means $\frac{s_{2}^{t}}{n_{2}^{t}} - \frac{s_{3}^{t}}{n_{3}^{t}}$ is large in this case, the exploration term in the 3rd arm is not strong enough to overturn the exploitation of the 2nd arm in such a small time window, which is different from the case of $r=16$.

Similar findings of the typical behaviors of the optimal arm leading to large regrets have also been reported before~\cite{Audibert2009, Fan2021}. Our analytical method complements these studies and provides a much more detailed picture of the dominant trajectories in this regime.

\subsubsection{Extremely unlucky events: Conditioned on $r=36$ and $r=40$}
\begin{figure}
    \centering
    \includegraphics[scale=1.05]{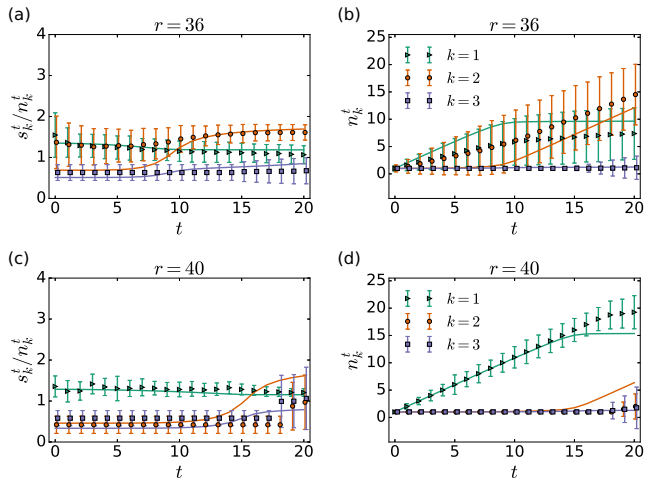}
    \caption{Dominant trajectories at regrets $r=36$ (panel (a) and (b)) and $r=40$ (panel (c) and (d)). The parameter setting is the same as those in Fig.~\ref{fig:traj_small_r}.}
    \label{fig:traj_big_r}
\end{figure}

In Fig.~\ref{fig:traj_big_r}, we consider the extremely unlucky cases with large regrets with $r=36$ and $r=40$, which enters the third convex branch of the potential $\Phi^{*}(r)$ shown in Fig.~\ref{fig:Phi_vs_r_vary_gamma_c}(c). The resulting phenomena are very similar to the cases in Sec.~\ref{sec:traj_r16_r24}, except that both the 3rd arm and the 2nd arm unluckily yield rewards smaller than the 1st arm (the worst arm). So the excessive exploitation of the worst arm leads to such high regrets being considered.

Lastly, we notice that for trajectories conditioned on $r=16$ and $r=36$, numerical simulations exhibit very large fluctuations. In these cases, there exist many trajectories having similar values of $\Phi^{*}$ as the optimal path (e.g., there can be multiple solutions of the saddle point equations with similar potentials). Our theory just picks the one with the smallest $\Phi^{*}$, the most probable trajectory, assuming a vanishing arm noise. However, when the noise is finite, different dominant and sub-dominant trajectories can coexist.

\subsection{Exact Results of a Small System}
In this section, we further investigate the nature of multiple solutions of the saddle point solution equations, which is related to the behavior of multiple branches of the stochastic action $\Phi^{*}(r|\sigma_{k})$. To this end, we consider a small MAB system with $K = 2, T = 1$, and set the arm-related parameters as $\boldsymbol{\mu} = [1, 2]^{\top}, \boldsymbol{\sigma} = \sqrt{\gamma}[1, 1]^{\top}$. For such a small system, the saddle point equations developed in Sec.~\ref{sec:saddle} and Sec.~\ref{sec:sim_saddle} can be straightforwardly reduced to the following self-consistent equation of a single variable $\Delta s^{0} = s_{2}^{0} - s_{1}^{0}$, 
\begin{align}
& g(\Delta s^{0}) = - 2 (\mu_{2} - \mu_{1}) \beta \gamma \mathcal{S}(\beta \Delta s^{0}) \big[ 1-\mathcal{S}(\beta \Delta s^{0}) \big] \mathrm{i}\hat{r}(\Delta s^{0})  \nonumber \\
& \qquad \qquad + (\mu_{2} - \mu_{1}) - \Delta s^{0} = 0, \label{eq:g_of_Ds} \\
& \mathrm{i}\hat{r}(\Delta s^{0})  = \frac{(\mu_{2} - \mu_{1}) \big[ \mathcal{S}(\beta \Delta s^{0})-2 \big] + r}{3 \gamma},
\end{align}
where we have defined the sigmoid function $\mathcal{S}(x) := 1/(1+e^{-x})$. 

\begin{figure}
    \centering
    \includegraphics[scale=1.05]{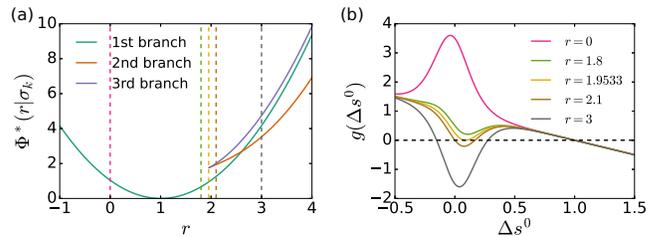}
    \caption{Behaviors of a small MAB with $K=2, T=1$. We set the parameters as $\sqrt{\gamma} = 0.4, \beta = 10$. (a) The stationary stochastic action $\Phi^{*}(r|\sigma_{k})$ determined by Eq.~(\ref{eq:g_of_Ds}). (b) The left-hand side of the self-consistent equation $g(\Delta s^{0})$ defined in Eq.~(\ref{eq:g_of_Ds}).}
    \label{fig:small_syst}
\end{figure}

The root $\Delta s^{0*}$ of the equation $g(\Delta s^{0})=0$ (for a given $r$) is a stationary point of $\Phi$. We set $\sqrt{\gamma}=0.4, \beta=10$ and identify all possible solutions for different regrets; the results are shown in Fig.~\ref{fig:small_syst}. For $r < r_{c} \approx 1.9533$, there is only one branch of solution with $\Delta s^{0*} \approx 1 (= \mu_{2} - \mu_{1})$; this corresponds to the ``liquid phase'' where the arm rewards deviate from their expected value homogeneously. For $r > r_{c} \approx 1.9533$, the system develops two additional branches of stationary solutions, where the optimal arm unluckily yields a small reward. The three branches of stationary solutions for large regrets correspond to two local minima (the 1st and 2nd branches in Fig.~\ref{fig:small_syst}) separated by a saddle point (the 3rd branch in Fig.~\ref{fig:small_syst}) in the stochastic action in the space of order parameters. In the small noise limit, the branch with the smallest potential $\Phi^{*}$ dominates the probability density. In the case of a finite noise strength, trajectories of different branches can coexist.

\section{Conclusion}
In summary, we employed the path-integral method from statistical physics to examine the complex dynamics of multi-armed bandits, which are classical prototypical models for understanding decision-making and optimization in uncertain environments and the dilemma between exploration and exploitation. 
By mapping the MABs under the UCB algorithm onto stochastic dynamical systems and considering the low temperature and small noise limit, we derived a large deviation formalism for the cumulative regrets of MABs, from which many valuable insights of the regret distribution and the dominant trajectory leading to a certain regret were obtained, revealing the transient dynamics leading lucky, unlucky or even extreme events. 

We observed that the stochastic action of the MABs can be non-convex, indicating the multimodal structure of the regret distribution. We showed that such a multimodal structure is due to the existence of multiple solutions of the saddle point equations when the regret is larger than a threshold. Under the UCB algorithm, the MABs have a much higher chance to be unlucky (the regret is higher than expected) than lucky (the regret is lower than expected), where the chance of being unlucky depends on the strength of exploration. The dominant trajectories leading to high regrets are those where the optimal arm unluckily yields a small reward in the initial exploration stage, such that it is much less exploited in the subsequent time steps.

Despite a relatively small time window being considered, our study uncovered many interesting characteristics of the MABs as complex systems. 
Extending the analysis to larger systems and longer time windows requires a sophisticated method to solve the high-dimensional nonlinear saddle point equations, which contributes to a more in-depth understanding of MABs and the exploitation-exploration dilemma, as well as the origin and occurrence of extremely unlucky events.
It would also be interesting to consider more complex reward distribution of the arms, such as the Gaussian mixture distribution.
We envisage that the methods developed in this work can provide a valuable tool for analyzing different variants of MABs and other complex stochastic decision-making processes.

\begin{acknowledgments}
We thank David Saad for the helpful discussions and the anonymous reviewers for their valuable suggestions. B.L. acknowledges support from the Leverhulme Trust (RPG-2018-092), European Union's Horizon 2020 research and innovation programme under the Marie Sk{\l}odowska-Curie Grant Agreement No. 835913, the startup funding from Harbin Institute of Technology, Shenzhen (Grant No. 20210134), and the National Natural Science Foundation of China (Grant No. 12205066).
C.H.Y. is supported by the Research Grants Council of the Hong Kong Special Administrative Region, China (Projects No. EdUHK GRF 18304316, No. GRF 18301217, and No. GRF 18301119), the Dean's Research Fund of the Faculty of Liberal Arts and Social Sciences (Projects No. FLASS/DRF 04418, No. FLASS/ROP 04396, and No. FLASS/DRF 04624), and the Internal Research Grant (Project No. RG67 2018-2019R R4015 and No. RG31 2020-2021R R4152), The Education University of Hong Kong, Hong Kong Special Administrative
Region, China.
\end{acknowledgments}

\section*{Author Declarations}
\subsection{Conflict of Interest}
The authors have no conflicts to disclose.

\subsection{Author Contributions}
\textbf{Bo Li:} Conceptualization (equal); Methodology (equal); Formal Analysis (lead); Software (lead); Writing – original draft (lead).
\textbf{Chi Ho Yeung:} Conceptualization (equal); Methodology (equal); Formal Analysis (supporting); Software (supporting); Writing – original draft (supporting).

\section*{Data Availability Statement}
Data sharing is not applicable to this article as no new data were created or analyzed in this study.

\appendix
\section{Details of the Path-integral Formalism}\label{sec:appendix_a}
In this section, we supplement some details of the path integral calculation. 

To perform integration over the disorder variables $x^{1:T}_{k}$, we first notice
\begin{align}
& \sum_{t=1}^{T}\hat{s}_{k}^{t}\left[\sum_{\tau=1}^{t}x_{k}^{\tau}\delta(a^{\tau},k)\right]	=\sum_{t=1}^{T}\hat{s}_{k}^{t}\left[\sum_{\tau=1}^{T}x_{k}^{\tau}\delta(a^{\tau},k)\mathbb{I}(t\geq\tau)\right] \nonumber \\
& = \sum_{\tau=1}^{T}x_{k}^{\tau}\delta(a^{\tau},k)\left[\sum_{t=1}^{T}\hat{s}_{k}^{t}\mathbb{I}(t\geq\tau)\right] \nonumber \\
& = \sum_{\tau=1}^{T}x_{k}^{\tau}\delta(a^{\tau},k)\left[\sum_{t=\tau}^{T}\hat{s}_{k}^{t}\right] = \sum_{t=1}^{T}x_{k}^{t}\delta(a^{t},k)\bigg[\sum_{\tau=t}^{T}\hat{s}_{k}^{\tau}\bigg],
\end{align}
where we have switched the dummy time indices $t$ and $\tau$ in the last line. 

Then for the terms relevant to $x^{1:T}_{k}$, we have
\begin{align}
& \int\prod_{t=1}^{T}\mathrm{d}x_{k}^{t}\mathcal{N}(x_{k}^{t}|\mu_{k},\sigma_{k}^{2})\exp\bigg\{\mathrm{i}\sum_{t=1}^{T}x_{k}^{t}\delta(a^{t},k)\bigg[\sum_{\tau=t}^{T}\hat{s}_{k}^{\tau}\bigg]\bigg\} \nonumber \\
& = \exp\bigg\{-\frac{1}{2}\sigma_{k}^{2}\sum_{t=1}^{T}\delta(a^{t},k)\bigg[\sum_{\tau,\tau'=t}^{T}\hat{s}_{k}^{\tau}\hat{s}_{k}^{\tau'}\bigg] \nonumber \\ 
& \qquad\qquad + \mathrm{i}\mu_{k}\sum_{t=1}^{T}\delta(a^{t},k)\bigg[\sum_{\tau=t}^{T}\hat{s}_{k}^{\tau}\bigg]\bigg\}, \label{eq:int_x}
\end{align}
where we have made use of the characteristic function of Gaussian distribution $\int e^{-\mathrm{i}xh}\mathcal{N}(x|\mu,\sigma^{2})dx=e^{-\frac{1}{2}\sigma^{2}h^{2}-\mathrm{i}\mu h}$ and the idempotence property of Kronecker delta $\delta(a^{t},k)^{2} = \delta(a^{t},k)$.

The first term in the exponent of Eq.~(\ref{eq:int_x}) can be rearranged as
\begin{align}
& \quad \sum_{t=1}^{T}\delta(a^{t},k)\bigg[\sum_{\tau,\tau'=t}^{T}\hat{s}_{k}^{\tau}\hat{s}_{k}^{\tau'}\bigg]	 \nonumber \\
& = \sum_{t=1}^{T}\delta(a^{t},k)\bigg[\sum_{\tau,\tau'=1}^{T}\hat{s}_{k}^{\tau}\hat{s}_{k}^{\tau'}\mathbb{I}(\tau\geq t)\mathbb{I}(\tau'\geq t)\bigg] \nonumber \\
& = \sum_{\tau,\tau'=1}^{T}\hat{s}_{k}^{\tau}\hat{s}_{k}^{\tau'}\bigg[\sum_{t=1}^{\min(\tau,\tau')}\delta(a^{t},k)\bigg] \nonumber \\
& = \sum_{\tau,\tau'=1}^{T}\hat{s}_{k}^{\tau}\hat{s}_{k}^{\tau'}\big(n_{k}^{\min(\tau,\tau')}-1\big),
\end{align}
where we have made use of the definition of $n_{k}^{t}$. 

Similarly, we have
\begin{align}
\sum_{t=1}^{T}\delta(a^{t},k)\bigg[\sum_{\tau=t}^{T}\hat{s}_{k}^{\tau}\bigg] & = \sum_{\tau=1}^{T}\hat{s}_{k}^{\tau}\bigg[\sum_{t=1}^{\tau}\delta(a^{t},k)\bigg] \nonumber \\
& = \sum_{\tau=1}^{T}\hat{s}_{k}^{\tau}\big(n_{k}^{\tau}-1\big).
\end{align}

Integration over $x_{k}^{0}$ for the relevant terms gives
\begin{align}
& \quad \int dx_{k}^{0}P(x_{k}^{0})\exp\bigg\{\mathrm{i}x_{k}^{0}\bigg[\sum_{\tau=1}^{T}\hat{s}_{k}^{\tau}\bigg]\bigg\} \nonumber \\ 
& = \exp\bigg\{-\frac{1}{2}\sigma_{k}^{2}\bigg[\sum_{\tau,\tau'=0}^{T}\hat{s}_{k}^{\tau}\hat{s}_{k}^{\tau'}\bigg]+\mathrm{i}\mu_{k}\bigg[\sum_{\tau=0}^{T}\hat{s}_{k}^{\tau}\bigg]\bigg\}.
\end{align}

Collecting the above results gives rise to the stochastic action defined in Eq.~(\ref{eq:Pr_of_Phi}) and Eq.~(\ref{eq:Phi_def}).

\section{Iteration Method for Solving the Saddle-point Equations} \label{sec:iteration_method}
Grouping all order parameters as a big vector $\boldsymbol{y} := [ \boldsymbol{s}^{0:T}, \boldsymbol{n}^{0:T}, \mathrm{i}\hat{\boldsymbol{s}}^{0:T}, \mathrm{i}\hat{\boldsymbol{n}}^{0:T} ]$, the saddle point equations have the form of 
\begin{align}
\boldsymbol{y} & = f(\boldsymbol{y}; \mathrm{i}\hat{r}), \label{eq:saddle_iter1} \\
\sum_{k} s_{k}^{T} & = (T+k)\mu_{*} - r, \label{eq:saddle_iter2}
\end{align}
where the nonlinear mapping $f(\cdot)$ is the right-hand side of the saddle point equations involving $\boldsymbol{y}$ given in Sec.~\ref{sec:saddle} and Sec.~\ref{sec:sim_saddle}. 

Given a random initial guess of the solution $[\boldsymbol{y}, \mathrm{i}\hat{r}]$, the iteration method iteratively performs the following two steps until convergence
\begin{enumerate}
    \item $\boldsymbol{y}^{\text{new}} = \alpha f(\boldsymbol{y}^{\text{old}};  \mathrm{i}\hat{r}^{\text{old}}) + (1-\alpha) \boldsymbol{y}^{\text{old}}$, 
    \item update $\mathrm{i}\hat{r}$ to bring $\sum_{k} s_{k}^{T}$ closer to $(T+k)\mu_{*} - r$.
\end{enumerate}
The second step is difficult to achieve as $s_{k}^{T}$ depends intricately on $\mathrm{i}\hat{r}$. In this step, we adopt a heuristic treatment of Eq.~(\ref{eq:skt_forward_1}) (for the final time $t=T$) as 
\begin{align}
s_{k}^{T}(\mathrm{i}\hat{r}) & = \sigma_{k}^{2}(\mathrm{i}\hat{s}_{k}^{T})n_{k}^{T} +     \mu_{k}n_{k}^{T} + \sigma_{k}^{2}\sum_{t'=0}^{T-1}(\mathrm{i}\hat{s}_{k}^{t'})n_{k}^{t'}, \\
& = - (\mathrm{i}\hat{r}) \sigma_{k}^{2} n_{k}^{T} + \text{const w.r.t. } \mathrm{i}\hat{r},
\end{align}
which we only retain the dependence of $\mathrm{i}\hat{s}_{k}^{T}$ on $\mathrm{i}\hat{r}$ (remind that $\mathrm{i}\hat{s}_{k}^{T} = - \mathrm{i}\hat{r}$) and consider the dependence of $n_{k}^{t}$ and $\mathrm{i}\hat{s}_{k}^{t'}$ for $t'<T$ on $\mathrm{i}\hat{r}$ as a weaker effect.

The above iteration method does not converge in some problem instances of MABs, in which cases it still leads to a significant reduction of the residual 
\begin{equation}
\mathrm{res} = ||\boldsymbol{y} - f(\boldsymbol{y};\mathrm{i}\hat{r})||^2 + \big( \sum_{k} s_{k}^{T} + r - (T+k)\mu_{*} \big)^2.    
\end{equation}
In these cases, we use the final outcome of the iteration method as an initial guess of the solution, which is fed to a nonlinear solver (e.g., by Newton's method) to find the root of the saddle point equation.

\bibliographystyle{unsrt}
\bibliography{reference}

\end{document}